\documentclass[runningheads]{llncs}

\usepackage[T1]{fontenc}
\usepackage{graphicx, tabularx}
\usepackage{booktabs}
\usepackage{url}
\usepackage{cite}
\usepackage{algpseudocode}
\usepackage{algorithm}% http://ctan.org/pkg/algorithm
\usepackage[hidelinks]{hyperref}

\usepackage[dvipsnames]{xcolor}
\usepackage{color}

\begin{document}
\title{Cross-Task Attention Network: Improving Multi-Task Learning for Medical Imaging Applications}
\titlerunning{CTAN: Improving Multi-Task Learning for Medical Imaging Applications}
\author{Sangwook Kim\inst{1}\orcidID{0000-0001-6482-9561} \and
Thomas G. Purdie\inst{1, 2, 4, 8}\orcidID{0000-0003-4176-8457} \and
Chris McIntosh\inst{1, 2, 3, 5, 6, 7}\orcidID{0000-0003-1371-1250}}
\authorrunning{S. Kim et al.}
\institute{
Department of Medical Biophysics, University of Toronto, Toronto, Canada \and
Princess Margaret Cancer Centre, University Health Network, Toronto, Canada \and
Toronto General Research Institute, University Health Network, Toronto, Canada \and
Princess Margaret Research Institute, University Health Network, Toronto, Canada \and
Peter Munk Cardiac Centre, University Health Network, Toronto, Canada \and
Department of Medical Imaging, University of Toronto, Toronto, Canada \and
Vector Institute, Toronto, Canada \and
Department of Radiation Oncology, University of Toronto, Toronto, Canada
\email{\{sangwook.kim,tom.purdie,chris.mcintosh\}@rmp.uhn.ca}}

\maketitle

\newcommand{\CM}[1]{{\footnote {\color{red}CM: #1}}}
\newcommand\figref{Fig.~\ref}

\begin{abstract}
Multi-task learning (MTL) is a powerful approach in deep learning that leverages the information from multiple tasks during training to improve model performance. In medical imaging, MTL has shown great potential to solve various tasks. However, existing MTL architectures in medical imaging are limited in sharing information across tasks, reducing the potential performance improvements of MTL. In this study, we introduce a novel attention-based MTL framework to better leverage inter-task interactions for various tasks from pixel-level to image-level predictions. Specifically, we propose a Cross-Task Attention Network (CTAN) which utilizes cross-task attention mechanisms to incorporate information by interacting across tasks.
We validated CTAN on four medical imaging datasets that span different domains and tasks including: radiation treatment planning prediction using planning CT images of two different target cancers (Prostate, OpenKBP); pigmented skin lesion segmentation and diagnosis using dermatoscopic images (HAM10000); and COVID-19 diagnosis and severity prediction using chest CT scans (STOIC). Our study demonstrates the effectiveness of CTAN in improving the accuracy of medical imaging tasks. Compared to standard single-task learning (STL), CTAN demonstrated a 4.67\% improvement in performance and outperformed both widely used MTL baselines: hard parameter sharing (HPS) with an average performance improvement of 3.22\%; and multi-task attention network (MTAN) with a relative decrease of 5.38\%. These findings highlight the significance of our proposed MTL framework in solving medical imaging tasks and its potential to improve their accuracy across domains.

\keywords{Multi-Task Learning  \and Cross Attention \and Automated Radiotherapy}
\end{abstract}

\section{Introduction}
\begin{figure}[!ht]
    \centering
    \includegraphics[width=\textwidth]{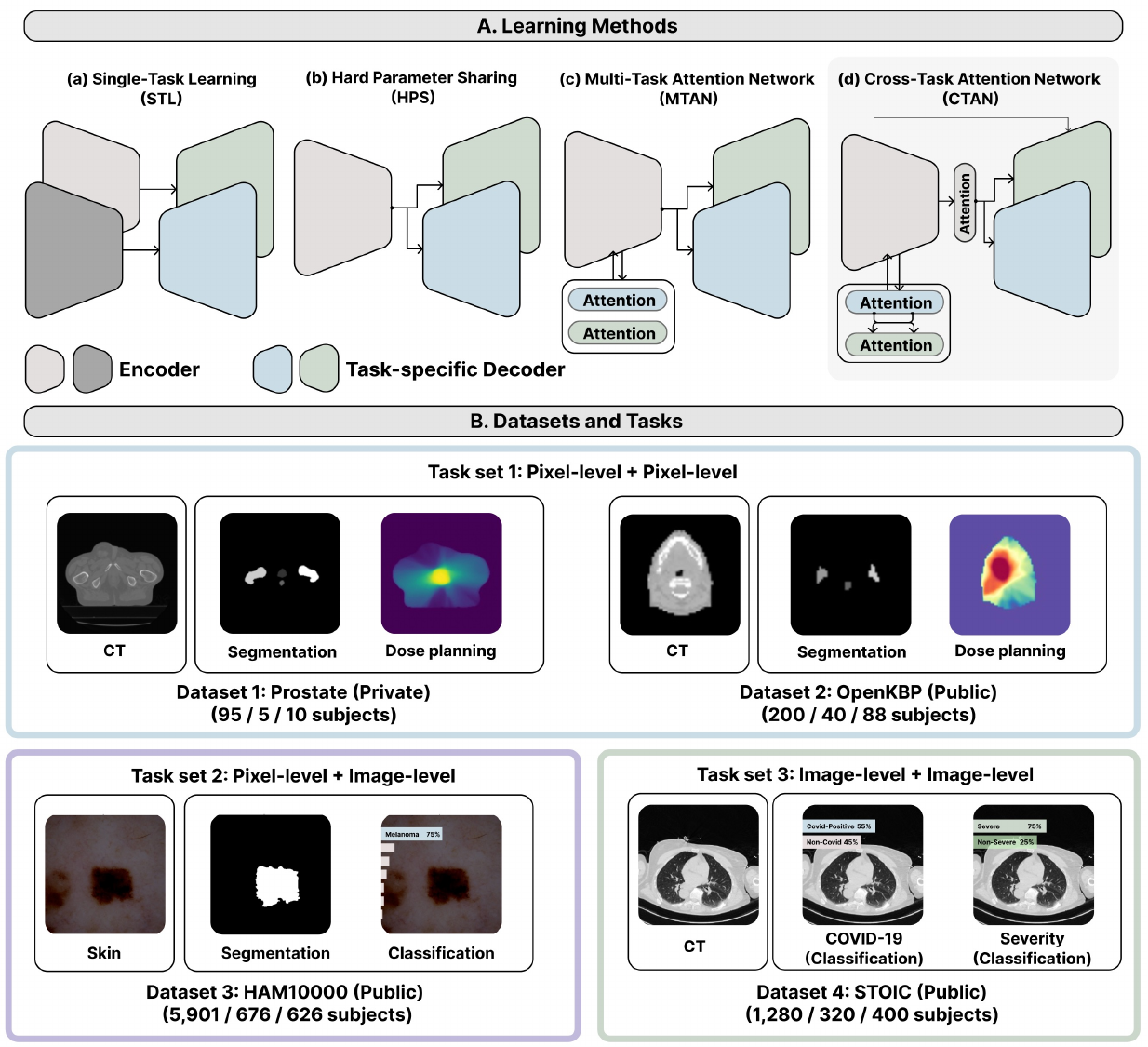}
    \caption{(Top) Cross-task attention network (CTAN) and other MTL model architectures: hard parameter sharing (HPS) \cite{andrychowicz2016learningHPS} and multi-task learning network (MTAN) \cite{liu2019endmtan}. Similar to the concept of one-to-many mappings from HPS and MTAN, CTAN has one shared encoder linked with decoders for each task. MTAN uses encoder features using attention for respective tasks. However, CTAN uses cross-attention in encoder and bottleneck layers to transfer task-specific features to task-specific decoders for better task interaction. (Bottom) Summary of four medical imaging datasets with three different task sets used in this study. \textbf{The number of samples of each train, validation, test splits are shown below each dataset.} Test datasets without complete segmentation labels and clinical information were excluded from the original datasets in OpenKBP and HAM10000, respectively.}
    \label{fig:datasetCTAN}
\end{figure}
Multi-task learning (MTL) \cite{caruana1998multitask} algorithms train deep learning models for two or more tasks simultaneously using shared parameters between models to encourage beneficial cooperation. MTL provides additional information not by explicitly adding more datasets for model training but by implicitly extracting training signals from multiple related tasks from the existing dataset. The various tasks are thought to regularize shared components of the network, leading to improved model performance and generalization. For example, following \cite{ashraf2022melanoma}, it is natural to assume that learning features required to delineate a skin lesion from the background may be relevant in comparing the lesion to its surrounding areas to inform the diagnosis.

Previous studies have demonstrated that learning two relevant tasks can improve model performance using MTL in medical imaging \cite{wimmer2021multi-mia05, chen2019multi-mia02, chen2019multi-mia04, boutillon2021multi-mia01, sainz2020multi-mia03, weninger2020multi}. Sainz et al., show the application and improvement of the model performance using MTL in breast cancer screening by training classification and detection of abnormal mammography findings \cite{sainz2020multi-mia03}. Chen et al., utilize MTL to improve atrial segmentation and classification using MRI\cite{chen2019multi-mia04}. Weninger et al., propose an MTL framework to improve brain tumour segmentation by jointly training detection of enhancing tumour and image reconstruction using brain MRI\cite{weninger2020multi}.

These studies demonstrate the applicability of MTL to improve performance for tasks in medical imaging. However, even though these studies have shown enhanced performance using MTL, most MTL architectures are based on hard-parameter sharing (HPS) \cite{andrychowicz2016learningHPS}, which includes a single shared encoder with task-specific decoders in a one-to-many fashion, maximizing encoder regularization between tasks but limiting all tasks to an identical feature set as opposed to some common features.

Introduced by Liu et al., multi-task attention network (MTAN) \cite{liu2019endmtan} also employs a one-to-many mapping but adds task-specific independent attention mechanisms that, while they can change the features of the embedding per task, they are not themselves able to share any information. With the introduction of MTAN, there have been studies using attention in MTL for automating binding between task features within the network architectures\cite{zhang2021survey, lopes2023cross}. However, most existing MTL studies using non-medical images focus on scenarios where all tasks are at the pixel-level. This is often impractical in the medical imaging domain, since acquiring pixel-level labels in medical images is impractical and labour-intensive. Thus, we focus on solving multi-task learning in hybrid scenarios including both pixel and image-level tasks by utilizing cross-task attention in MTL using medical imaging datasets.

We hypothesize that by leveraging the shared feature abilities of HPS with the flexibility of MTAN through a novel cross-task attention framework that shares task information across the attention mechanisms, we can better utilize inter-task interaction to improve overall performance using MTL. Additionally, cross-attention of bottleneck features for each task was also employed to provide cross-task dependent information to decoders for each task. We validated our approach using three distinct pairs of tasks from four medical imaging datasets. CTAN shows broad applicability with mixes of tasks at the both the pixel and image-level.

\subsubsection{Contributions}
\begin{figure}[!ht]
    \centering
    \includegraphics[width=\textwidth]{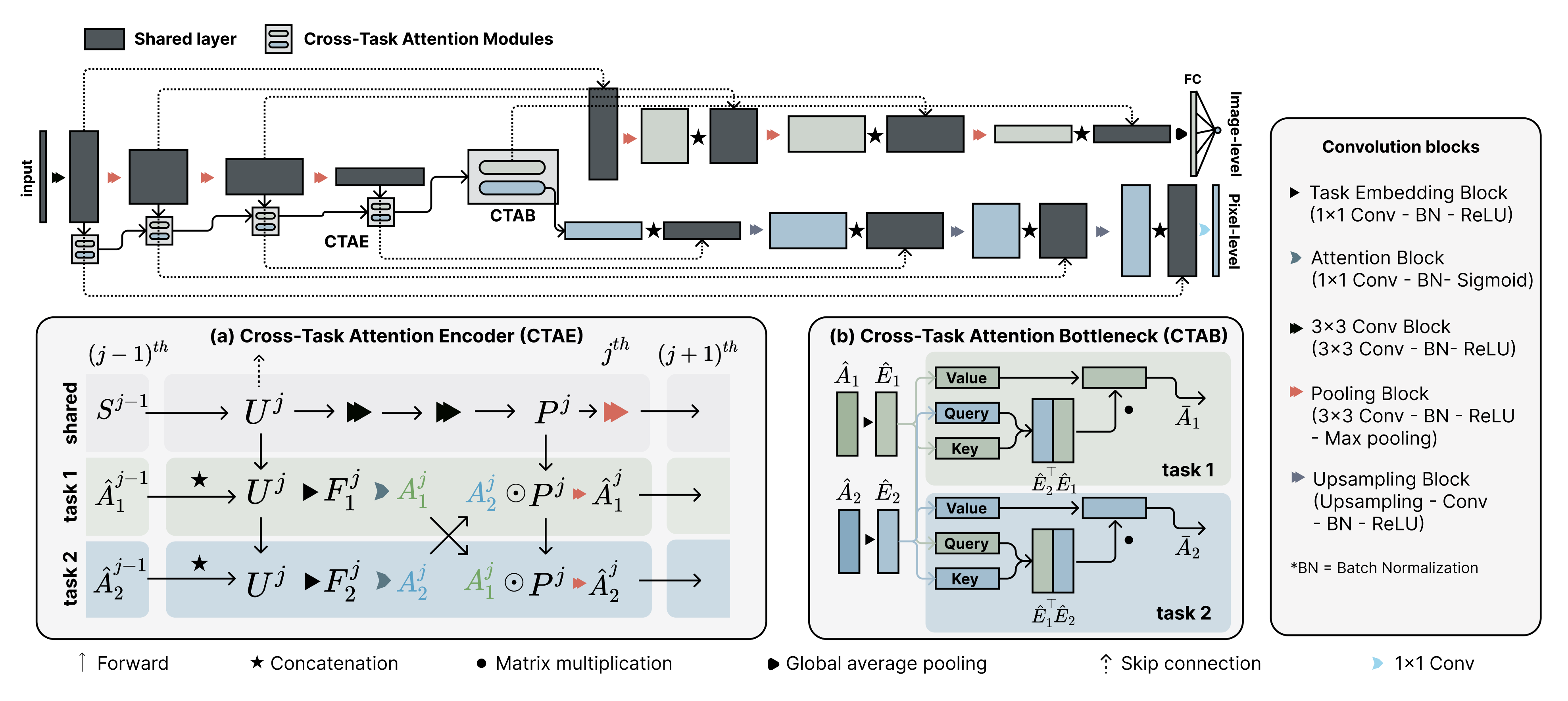}
    \caption{Overview of architecture of cross-task attention network (CTAN), including the encoder and two decoders for image-level and pixel-level tasks. Convolution blocks are shown on the right, along with the two cross-task attention modules: (a) Cross-task attention encoder (CTAE), and (b) Cross-task attention bottleneck (CTAB).}
    \label{fig:CTAECTAB}
\end{figure}

We propose a novel Cross-Task Attention Network (CTAN), an MTL framework that leverages cross-task attention modules in the encoder and bottleneck layer to capture inter-task interaction across tasks (see \figref{fig:CTAECTAB}). Our results demonstrate that CTAN is effective in learning three types of vision tasks, including two pixel-level prediction tasks and one image-level task from various domains. As shown in \figref{fig:datasetCTAN}, we experimented with three different task pairs from four datasets. In addition, we showed the performance improvement of CTAN compared to single-task learning (STL), and two widely used MTL baseline architectures, HPS and MTAN. 

\section{Methods and Materials}
\subsection{Cross-Task Attention Network (CTAN)}
CTAN consists of two cross-task attention modules, the cross-task attention encoder (CTAE), and the cross-task attention bottleneck (CTAB) (see Fig. \ref{fig:CTAECTAB}).
CTAE is employed within the encoder layers by calculating the attentive mask, and uses two pieces of information targeted for each task. CTAE enables the encoder to extract task-specific information in the encoder. It encodes and decodes the input features to highlight and extract significant features. The attention module in CTAE resembles the attention module in \cite{liu2019endmtan}, wherein for each task Liu et al. calculate attention maps using one attention block per task and multiply with the feature maps during a forward pass with data from that task. However, in CTAE, attention maps are instead multiplied in a cross-direction way, as shown in Fig. \ref{fig:CTAECTAB}-a. This helps the model to integrate the shared features by multiplying the cross-task attentive maps with features from the shared block, which enables an inter-task interaction while training. We denote $U^j$ and $P^j$ as features from $j^{th}$ layer of the shared encoder, and $t$ as task index. Note that $P^j$ refers to the output of two convolution blocks using $U^j$ as the input. $S^{j-1}$ denotes the input of $j^{th}$ layer in the shared encoder, which is the output of the shared block in $j-1^{th}$ layer for $j>1$. Whereas, when $j=1$, the input image embedding from the 3x3 Conv block is used (see \figref{fig:CTAECTAB}). The task-specific embedded features, $F^j_t$, result from the concatenation of $U^j$ and $\hat{A}^{j-1}_t$ for $j>1$, while $U^j$ for $j=0$, followed by the task embedding block in \figref{fig:CTAECTAB}. $F^j_t$ is then fed into the task-specific attention block to create attention mask $A^j_t$. The output of CTAE $\hat{A}^j_t$ is defined as:
\begin{equation}
    \hat{A}^j_t= Pool(A^j_{t'}\ \odot\ P^j),\ t \in \{\it{1}, \it{2}\}, \\
\end{equation}
where $Pool$ refers to the pooling block (see \figref{fig:CTAECTAB}), $\odot$ refers to the element-wise multiplication, and $t'$ refers to the task index of the other task trained together. $\hat{A}^j_t$ then serves as the input attention mask for the attention block in the next layer, propagating attention across the decoder ($\hat{A}^{j-1}_t$ is set to all zero for the first layer). 

We propose CTAB as shown in \figref{fig:CTAECTAB}-b, in which we calculate and multiply cross-task attention of two task-embedded features to task-specific bottleneck representation. We calculate the cross-task attention mask using a $query$ and a $key$ and apply the attention mask to a $value$. Herein, $value$ and $key$ are the same task-embedded features, and $query$ is the embedding of the other task. Thus, the output of CTAB $\bar{A}_t$ is defined as:
\begin{equation}\label{eq:ctab}
    \bar{A}_t= \hat{E}_t\ \cdot (\hat{E}^\top_{t'} \cdot \hat{E}_t),\ t \in \{\it{1}, \it{2}\},
\end{equation}
where $\top$ refers to transpose of a matrix, $\cdot$ refers to matrix multiplication, and $\hat{E}_t$ denotes the task-specific embedded features for task $t$. The output of CTAB, $\bar{A}_t$, is forwarded to task-specific decoders. 

\subsubsection{Encoder and Decoder} 
We utilize a ResNet-50 \cite{he2016deep} pre-trained with ImageNet \cite{deng2009imagenet} as the encoder backbone, with identical architecture across all experiments. However, we implement different decoders for image-level and pixel-level tasks. For pixel-level tasks such as segmentation and dose prediction, we incorporate skip connections \cite{ronneberger2015u} between the encoders and decoders, with three up-sampling blocks using bilinear interpolation (as depicted in \figref{fig:CTAECTAB}), followed by a 1x1 convolution layer with output channels equal to the number of segmentation labels, and a single channel for dose prediction. For image-level tasks, we use decoders with skip connections and four down-sampling layers, with a global average pooling layer \cite{he2015delving} and a fully-connected layer at the end. Notably, we introduce skip connections in the classifier to balance model training and address asymmetric decoder issues that arise when training MTL to solve both image-level and pixel-level tasks together. Finally, we use a fully-connected layer with a sigmoid activation function for binary classification (STOIC) and a softmax function for multi-class classification (HAM10000) as the final output layer.

\subsection{Training details}
We use Adam \cite{kingma2014adam} optimizer with the learning rate of $10^{-4}$ and the weight decay of $10^{-5}$. We use task-specific losses (see Table \ref{tab:loss}). Dynamic Weight Averaging \cite{liu2019endmtan} was utilized to stabilize the combined training losses of all tasks. Batch size of 32 was used for the Prostate dataset, and 8 for the rest. We conducted experiments using PyTorch (ver 1.9.0) \cite{paszke2019pytorch}, with an NVIDIA A100 GPU with 40GB memory.

\begin{table}[ht!]
  \centering
  \caption{Summary of loss functions for each task. We use combo loss \cite{ma2021loss}, with the 0.3 and 0.7 for the loss weight of dice loss and cross-entropy loss, respectively.}
    \begin{tabularx}{0.99\linewidth}{p{2.0cm}p{8cm}X}
    \toprule
    Task & Loss function & Dataset \\
    \midrule
    Segmentation&Combo Loss\cite{ma2021loss}\newline (Weighted combination of Dice Loss and Cross-entropy Loss) & Prostate, OpenKBP, \newline HAM10000\\
    \midrule
    Dose \newline prediction&Mean absolute error (MAE) Loss\cite{openkbp}&Prostate, OpenKBP\\
    \midrule
    Classification&Cross-entropy Loss&HAM10000, \newline STOIC\\
    \bottomrule
    \end{tabularx}
  \label{tab:loss}
\end{table}

\subsection{Evaluation}\label{sec:eval}
We used task-specific metrics to evaluate the model performance for each task: dice similarity coefficient for segmentation(\%); mean absolute error(Gy) between ground truth and predicted dose distribution maps for dose prediction; accuracy(\%) for classification of HAM10000; and the area under the receiver operating characteristic curve(\%) for classification of STOIC. Following \cite{liu2021conflict}, we define the relative performance of MTL models compared to STL:
\begin{equation}
    % \Delta_{mean}(\%) = \frac{100}{2}\sum_{i=1}^{2}{\frac{(-1)^{l_i}(M_{b,i} - M_{m,i})}{M_{b,i}}},\ l \in \{0, 1\},
    \Delta_{task}(\%) = 100 * \frac{(-1)^{l_i}(M_{b,i} - M_{m,i})}{M_{b,i}},\ l \in \{\it{0}, \it{1}\},
\end{equation}
where $i$ denotes the index of the task, $m$ and $b$ refer to the target MTL model and the baseline STL, respectively. $M$ refers to the task performance metric. $l$ denotes the metric-specific flag, where $1$ if the metric is higher the better, and vice versa. We can then calculate the average of the relative difference of all task-specific metrics for each experiment. Positive value of relative performance represents the performance of MTL is better than that of STL. 

\subsection{Datasets}
We validated our approach using four medical imaging datasets with three different task sets (see Fig. \ref{fig:datasetCTAN}-B). The first task set consists of two pixel-level tasks: dose prediction and segmentation of organs at risk (OAR) and clinical target volume (CTV) for prostate (Prostate) and head and neck cancer treatment (OpenKBP )(\url{https://www.aapm.org/GrandChallenge/OpenKBP}, \cite{openkbp}). Segmentation labels for the Prostate dataset are rectum, bladder, left and right femur, while brain stem, spinal cord, left and right parotid are used in OpenKBP. Patients For the second task set, which contains one image-level and one pixel-level tasks, dermatoscopic images of pigmented skin lesion datasets (HAM10000) (\url{https://doi.org/10.7910/DVN/DBW86T}, \cite{ham10000}) are used to segment and diagnose skin lesions. The last set has two image-level tasks: classification of COVID-19 and disease severity using chest CT scans (STOIC) (\url{https://stoic2021.grand-challenge.org}, \cite{stoic}). 
\section{Experiments and Results}
In Table \ref{tab:result}, the results showed that CTAN outperformed STL with an average relative difference of 4.67\%. For the Prostate and OpenKBP datasets, which have two different pixel-level tasks, CTAN showed an improvement of 2.18\% and 1.99\%, respectively, over STL. In both datasets, the performance increase for dose prediction task was larger than that of segmentation task. Notably, CTAN improved the performance of dose prediction when the task is trained with segmentation of organs at risk and target volumes, rather than improving the performance of segmentation. For HAM10000, CTAN showed an overall performance improvement with a significant increase in diagnosing skin lesions. However, the performance of segmenting pigmented lesions marginally improved compared to the classification task. For STOIC, CTAN resulted in an average relative difference of 4.67\% for both image-level tasks, with a significant increase in diagnosing severe cases but a decrease in diagnosing COVID-19.

As shown in Table \ref{tab:result}, CTAN outperformed both HPS and MTAN with an average relative improvement of 3.22\% and relative decrease of 5.38\%, compared to STL, respectively. Unlike other MTL baselines, CTAN showed performance improvement regardless of task groups combined with different task-levels. However, there were cases where CTAN did not outperform other baselines at the single task level. For instance, for the Prostate datasets' segmentation task, HPS outperformed CTAN with a relative difference of 1.74\% while CTAN showed only a 0.54\% increase. Nevertheless, overall performance gain using CTAN was higher across datasets and tasks, indicating that the cross-task attention mechanisms in CTAN were effective in learning multiple tasks.

\begin{table}[ht!!!]
  \centering
  \caption{Results of task-specific metrics ($M_{task}$) and their relative difference to STL ($\Delta_{task}$) of STL, HPS, MTAN, and CTAN on four datasets. Higher values are the better for all metrics, except for $M_{task 2}$ in the Prostate and OpenKBP datasets. Best and second-best results are bolded and underlined, respectively. Average values are only calculated for relative performance difference of MTL methods.}
    \begin{tabularx}{0.99\linewidth}{lp{1.5cm}p{1.1cm}p{1.6cm}p{1.0cm}p{1.6cm}p{1.4cm}c}
    \toprule
    Dataset & Method & {$M_{task 1}$} & {\textbf{$\Delta_{task1}\uparrow$}} & {$M_{task 2}$} & {\textbf{$\Delta_{task2}\uparrow$}} & $\Delta_{mean}\uparrow$ & Rank\\
    \midrule
    Prostate & STL   &81.96 &  &0.93&  &  & 3 \\
    & HPS   &\textbf{83.28} & \textbf{1.74\%} &\underline{0.91}& \underline{1.29\%} & \underline{1.51\%} &2 \\
          & MTAN  &75.47& -7.92\% &0.99 & -7.29\% & -7.60\% & 4 \\
          & \textbf{CTAN} &\underline{82.40}& \underline{0.54\%} &\textbf{0.89}& \textbf{3.82\%} & \textbf{2.18\%}&\textbf{1}\\
    \midrule
    OpenKBP\cite{openkbp} & STL   &\underline{71.29}&  &\underline{0.53}&  &  &2\\
    & HPS   &70.87& -0.52\% & \underline{0.53}& \underline{0.31\%} & \underline{-0.10\%} &3\\
          & MTAN  &66.09& -7.30\% &0.56& -5.29\% & -6.29\% &4\\
          & \textbf{CTAN} &\textbf{71.59}& \textbf{0.42\%} &\textbf{0.51}& \textbf{3.56\%} & \textbf{1.99\%} &\textbf{1}\\
    \midrule
    HAM10000\cite{ham10000} & STL   &\underline{92.83}&  &49.24&  &  &3\\
    &HPS   &92.21& -0.68\% &\underline{55.49}& \underline{12.69\%} & \underline{6.01\%} &2\\
          & MTAN  &92.15& -0.73\% &47.08& -4.37\% & -2.55\% &4\\
          & \textbf{CTAN} &\textbf{92.91}& \textbf{0.09\%} &\textbf{57.85}& \textbf{17.49\%} & \textbf{8.79\%} &\textbf{1}\\
    \midrule
    STOIC\cite{stoic} & STL & \textbf{71.88} &  &55.83&  &  &3\\
    & HPS & 63.84 & -11.18\% &\textbf{68.17}& \textbf{22.09\%} & \underline{5.45\%} &2\\
          & MTAN  &57.55& -19.93\% &61.30& 9.79\% & -5.07\% &4\\
          & \textbf{CTAN} &\underline{68.73}& \underline{-4.38\%} &\underline{64.66}& \underline{15.81\%} & \textbf{5.72\%} &\textbf{1}\\
    \midrule
    Average & STL &  &  & - &  &  &3\\
    & HPS   &\ \ \ -& \underline{-2.66}\% &\ \ \ -& \underline{9.09}\% & \underline{3.22}\% &2\\
          & MTAN  &\ \ \ -& -8.97\% &\ \ \ -& -1.79\% & -5.38\% &4\\
          & \textbf{CTAN} &\ \ \ -& \textbf{-0.83}\% &\ \ \ -& \textbf{10.17}\% & \textbf{4.67}\% &\textbf{1}\\
    \bottomrule
    \end{tabularx}
  \label{tab:result}
\end{table}

\section{Discussion}\label{sec:discussion}
Our findings suggest that CTAN can improve the MTL performance across three distinct tasks from four distinct medical imaging datasets by 4.67\% on average. However, the specific performance improvements on each dataset and task can vary. Compared to other tasks, CTAN only marginally improve performance in segmentation task.
This might be due to the faster convergence of segmentation tasks in comparison to others, which may cause them to act more as regularizers with pixel-level prior knowledge providing local contextual information for other tasks \cite{pinheiro2015imagelevel}. In this regard, results show that CTAN is more effective in utilizing segmentation tasks for learning high-level semantic cues compared to other MTL baselines. In particular, CTAN can implicitly learn to avoid dose exposure to OARs and maximize dose to the CTV by training two clinically relevant tasks. This implies a potential to automate dose planning without the dependence on the contouring information, prior to predicting the dose distribution. This approach can ensure robustness against the variability of human annotators and improve automated planning quality for clinical care \cite{mcintosh2021clinicalnatmed}. 

We observed a performance drop in COVID-19 classification in STOIC due to the intricate nature of the task, as diagnosing severity depends on the COVID-19 diagnosis and causes per-task gradient collision during training. However, CTAN proved to be effective in minimizing the performance drop in COVID-19 classification compared to other MTL methods. This implies CTAN can selectively learn cross-task attentive features to improve overall performance. Future work could expand the applications of CTAN to other domains such as videos of natural teeth \cite{katsaros2022multidental}, fundus photography for diagnosing glaucoma \cite{fang2022refuge2}, or laparoscopic hysterectomy \cite{wang2022autolaparo}, and further investigate what drives the per dataset variations.

In conclusion, we introduce a novel MTL framework, CTAN, that utilizes cross-task attention to improve MTL performance in medical imaging from multiple levels of tasks by 4.67\% compared to STL. Results demonstrate that incorporating inter-task interaction in CTAN enhances overall performance of three medical imaging task sets from four distinct datasets, surpassing STL and two widely-used baseline MTL methods. This highlights CTAN's effectiveness and potential to improve MTL performance in the field of medical imaging.

\bibliographystyle{splncs04}
\bibliography{reference}
\clearpage
\appendix
\setcounter{page}{1}

\begin{algorithm}[ht!!!]
  % \caption{PyTorch-like Pseudocode for Cross-task attention network (CTAN)}
  \caption{Pseudocode for CTAN and Weighted Combined Loss}
  \begin{algorithmic}[1]
    \Function{Cross-Task Attention Encoder (CTAE)}{$U^j$, $P^j$, $\hat{A}^{j-1}_{t}$, $\hat{A}^{j-1}_{t'}$}
        \State $F^j_t \gets TaskEmbeddingBlock(Concat(U^j, \hat{A}^{j-1}_t))$ \Comment{Task embedding for task $t$}
        \State $F^j_{t'} \gets TaskEmbeddingBlock(Concat(U^j, \hat{A}^{j-1}_{t'}))$ \Comment{Task embedding for task $t'$}
        % \State $\textcolor{YellowGreen}{A^j_t}, \textcolor{CornflowerBlue}{A^j_{t'}} \gets AttentionBlock(F^j_t), AttentionBlock(F^j_{t'})$\Comment{Attention for each task feature}
        \State $\textcolor{YellowGreen}{A^j_t} \gets AttentionBlock(F^j_t) $\Comment{Attention for each task-embedded feature}
        \State $\textcolor{CornflowerBlue}{A^j_{t'}} \gets AttentionBlock(F^j_{t'})$
        \State $\textcolor{YellowGreen}{\hat{A}^j_t} \gets PoolingBlock(Concat(P^j, \textcolor{CornflowerBlue}{A^j_{t'}}))$\Comment{\textbf{Cross-task attention}}
        \State $\textcolor{CornflowerBlue}{\hat{A}^j_{t'}} \gets PoolingBlock(Concat(P^j, \textcolor{YellowGreen}{A^j_t}))$ 
        \State \textbf{return} $\hat{A}^j_t, \hat{A}^j_{t'}$ \Comment{Return CTAE outputs for each task}
    \EndFunction
    \Function{Cross-Task Attention Bottleneck (CTAB)}{$\hat{A}_t$, $\hat{A}_{t'}$}
        \State $\textcolor{YellowGreen}{\hat{E}_t}, \textcolor{CornflowerBlue}{\hat{E}_{t'}} \gets TaskEmbeddingBlock(\textcolor{YellowGreen}{\hat{A}_t}), TaskEmbeddingBlock(\textcolor{CornflowerBlue}{\hat{A}_{t'}})$ 
        
        \State $\textcolor{YellowGreen}{\bar{A}_t} \gets \textcolor{YellowGreen}{\hat{E}_t} \cdot (\textcolor{CornflowerBlue}{\hat{E}^{\top}_{t'}} \cdot \textcolor{YellowGreen}{\hat{E}_t})$ \Comment{\textbf{Cross-task attention} of two task embedded features}
        
        \State $\textcolor{CornflowerBlue}{\bar{A}_{t'}} \gets \textcolor{CornflowerBlue}{\hat{E}_{t'}} \cdot (\textcolor{YellowGreen}{\hat{E}^{\top}_t} \cdot \textcolor{CornflowerBlue}{\hat{E}_{t'}})$
        
        \State \textbf{return} $\bar{A}_t, \bar{A}_{t'}$\Comment{Return CTAB outputs for each task}
    \EndFunction
    \Function{Cross-Task Attention Network (CTAN)}{$x$}
        \For{$j\gets0, n$}\Comment{Iterate through the $n$ encoder blocks}
            \If{$j = 0$}\Comment{For the $1^{st}$ encoder block}
                \State $S^{j-1} \gets ConvBlock(x)$ \Comment{Outputs of Conv block with input image $x$}
                \State $\hat{A}^{j-1}_t, \hat{A}^{j-1}_{t'} \gets 0, 0$\Comment{Prior attention information with zeros}
            \Else
                \State $S^{j-1} \gets PoolingBlock(P^{j-1})$ \Comment{Pooled features from the previous block}
            \EndIf
            \State $U^j \gets S^{j-1}$ \Comment{Feature maps from the shared block}
            \State $P^j \gets ConvBlock(ConvBlock(U^j))$\Comment{Outputs of two Conv blocks from $U^j$}
            \State $\hat{A}^{j}_t, \hat{A}^{j}_{t'} \gets CTAE(U^j, P^j, \hat{A}^{j-1}_{t}, \hat{A}^{j-1}_{t'})$\Comment{Cross-attention of encoder features}
        \EndFor
        \State $\bar{A}_t, \bar{A}_{t'} \gets CTAB(\hat{A}_t, \hat{A}_{t'})$\Comment{Cross-attention of bottleneck features}
        \State $y^{pred}_t, y^{pred}_{t'} \gets D_t(\bar{A}_t, U^{j\in[0,n]}), D_{t'}(\bar{A}_{t'}, U^{j\in[0,n]})$\Comment{Task-specific decoder $D_t$}
        \State \textbf{return} $y^{pred}_t, y^{pred}_{t'}$ \Comment{Return predicted outputs for each task}
    \EndFunction
    \Function{Weighted Combined Loss}{$x$, $\mathcal{L}^{e-1}_{t}$, $\mathcal{L}^{e-1}_{t'}$, $\mathcal{L}^{e-2}_{t}$, $\mathcal{L}^{e-2}_{t'}$}
        \If{$e < 2$}
            \State $\lambda^{e}_{t}, \lambda^{e}_{t'} \gets 1, 1$\Comment{Equal weighting for the $1^{st}$ and $2^{nd}$ iteration}
        \Else
            \State $\lambda^{e}_{t}, \lambda^{e}_{t'} \gets DWA(\mathcal{L}^{e-1}_{t}, \mathcal{L}^{e-1}_{t'}, \mathcal{L}^{e-2}_{t}, \mathcal{L}^{e-2}_{t'})$ \Comment{Loss weights using DWA\cite{liu2019endmtan}}
        \EndIf
        % \State $\lambda^{e}_{t'} \gets DWA(\mathcal{L}^i_{t'}, \mathcal{L}^{i-1}_{t'}, \mathcal{L}^{i-2}_{t'})$
        \State $y^{pred}_{t}, y^{pred}_{t'} \gets CTAN(x)$ \Comment{Outputs from CTAN for each task}
        % \State $\mathcal{L}^e_{t} \gets L_{t}(y^{pred}_{t}, y^{gt}_{t})$\Comment{Loss for task $t$}
        % \State $\mathcal{L}^e_{t'} \gets L_{t'}(y^{pred}_{t'}, y^{gt}_{t'})$\Comment{Loss for task $t'$}
        \State $\mathcal{L}^e_{t}, \mathcal{L}^e_{t'} \gets L_{t}(y^{pred}_{t}, y^{gt}_{t}), L_{t'}(y^{pred}_{t'}, y^{gt}_{t'})$\Comment{Task-specific loss function $L_t$}
        \State $\mathcal{L}^e_{total} = \lambda^{e}_{t}\mathcal{L}^e_{t} + \lambda^{e}_{t'}\mathcal{L}^e_{t'}$ \Comment{Weighted sum of the two losses}
        \State \textbf{return} $\mathcal{L}^e_{total}$\Comment{Return weighted combined loss}
    \EndFunction
  \end{algorithmic}
  \label{alg:ctan}
\end{algorithm}

We describe the pseudocode of cross-task attention network (CTAN) in Alg. \ref{alg:ctan}. The notations are equal to the main paper (see Sect 2.1 and Fig. 2). $Concat$ denotes a channel-wise concatenation of given inputs. $U^{j\in[0,n]}$ refers to the shared features, that are transferred from the shared encoder to decoders by skip connection. We use the weighted combination of task-specific losses using dynamic weight average (DWA) \cite{liu2019endmtan}. $\lambda^{e}_{t}$ refers to the loss weighting for task $t$ at epoch $e$. 
% When $e$ < 2,  both $\lambda^{e}_{t}$ and $\lambda^{e}_{t'}$ are equal to 1. After the second iteration ($e \geq$ 2), the loss weights are calculated using dynamic weight average (DWA) \cite{liu2019endmtan}.

% \bibliographystyle{splncs04}
% \bibliography{reference}

\end{document}